\documentclass{article}

\usepackage{algorithm}
\usepackage{algpseudocode}
\usepackage[numbers]{natbib}  
\usepackage[colorlinks=true, linkcolor=black, citecolor=blue, urlcolor=black]{hyperref}
\usepackage{wrapfig}
\usepackage{amsmath} 
\usepackage[preprint]{neurips_2025}

\usepackage{booktabs}             
\usepackage{multirow}             
\usepackage{graphicx}            
\usepackage{colortbl}       \usepackage{array}      
\usepackage{xcolor}    \usepackage{float}   \usepackage{subcaption}
\usepackage{enumitem}
\usepackage{bbold}
\definecolor{highlight}{RGB}{255, 199, 206}

\usepackage[utf8]{inputenc} % allow utf-8 input
\usepackage[T1]{fontenc}    % use 8-bit T1 fonts
\usepackage{hyperref}       % hyperlinks
\usepackage{url}            % simple URL typesetting
\usepackage{booktabs}       % professional-quality tables
\usepackage{amsfonts}       % blackboard math symbols
\usepackage{nicefrac}       % compact symbols for 1/2, etc.
\usepackage{microtype}      % microtypography
\usepackage{xcolor}         % colors

\title{ASTER: Adaptive Spatio-Temporal Early Decision Model for Dynamic Resource Allocation}

\author{%
}

\newcommand{\modelname}{\textnormal{ASTER}}

\begin{document}

\author{
Shulun Chen$^{1}$, Wei Shao$^{2}$, Flora D. Salim$^{1}$, Hao Xue$^{1}$ \\
$^1$University of New South Wales, Sydney, NSW, Australia \\
$^2$Data61, CSIRO, Clayton, Victoria, Australia \\
\texttt{shulun.chen@student.unsw.edu.au;}\texttt{phdweishao@gmail.com},\\ 
\texttt{\{flora.salim, hao.xue1\}@unsw.edu.au}
}
\maketitle

\begin{abstract}
Supporting decision-making has long been a central vision in the field of spatio-temporal intelligence. While prior work has improved the timeliness and accuracy of spatio-temporal forecasting, converting these forecasts into actionable strategies remains a key challenge. A main limitation is the decoupling of the prediction and the downstream decision phases, which can significantly degrade the downstream efficiency. For example, in emergency response, the priority is successful resource allocation and intervention, not just incident prediction. To this end, it is essential to propose an Adaptive Spatio-Temporal Early Decision model ($\modelname$) that reforms the forecasting paradigm from event anticipation to actionable decision support. This framework ensures that information is directly used for decision-making, thereby maximizing overall effectiveness. Specifically, $\modelname$ introduces a new Resource-aware Spatio-Temporal interaction module (RaST) that adaptively captures long- and short-term dependencies under dynamic resource conditions, producing context-aware spatiotemporal representations. To directly generate actionable decisions, we further design a Preference-oriented decision agent (Poda) based on multi-objective reinforcement learning, which transforms predictive signals into resource-efficient intervention strategies by deriving optimal actions under specific preferences and dynamic constraints. Experimental results on four benchmark datasets demonstrate the state-of-the-art performance of $\modelname$ in improving both early prediction accuracy and resource allocation outcomes across six downstream metrics. The code is available at \href{https://github.com/ccc5503/ASTER/}{\url{https://github.com/ccc5503/ASTER/}}.
\end{abstract}

\section{Introduction}
\label{s1}
Spatio-temporal prediction has become a pivotal technique for a broad range of domains, including traffic control, wildfire management, and crime prediction~\cite{joseph2019spatiotemporal,shao2022long,shao2024transferrable,xia2022spatial}. This approach models historical spatio-temporal patterns to predict future events. In response to the demand for prompt intervention in high-stakes scenarios, a growing body of research explores early forecasting~\cite{hartvigsen2020recurrent,li2017diffusion,shao2023early,shao2024stemo}, which is closely associated with emergency resource allocation in practice. For instance, in crime prevention, accurate but late forecasts lead to irreversible harm, while early but inaccurate alerts can trigger false alarms and deplete emergency resources, compromising future response. 

\begin{wrapfigure}{r}{0.50\textwidth}
  \centering
  \vspace{-10pt}
  \includegraphics[width=0.50\textwidth]{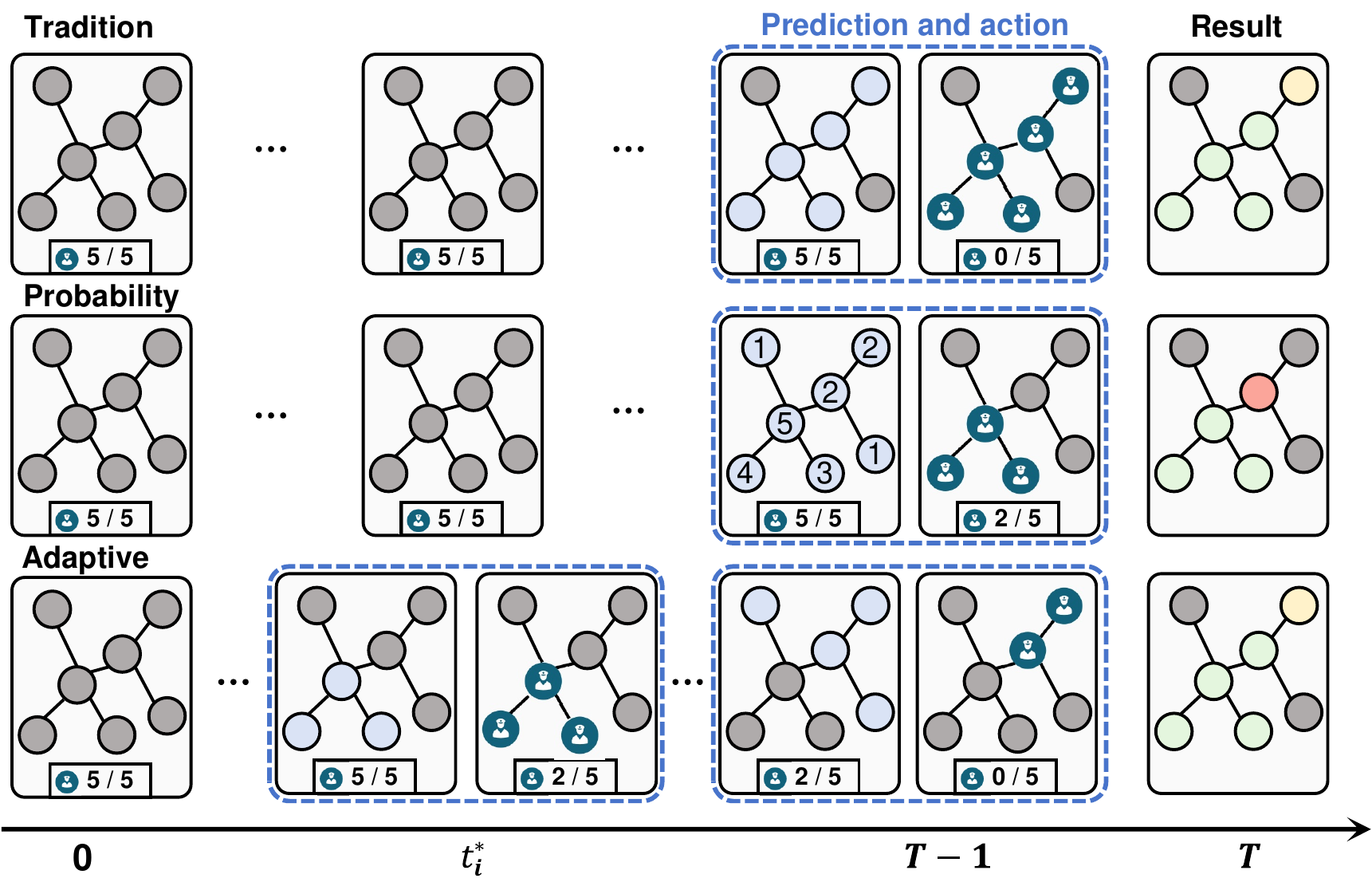}
  \vspace{-10pt}
  \caption{Example of three methods for early spatio-temporal decision-making. Each circle is a node (e.g., regions) with a recorded signal (e.g., accident occurrence) over time. In the result column, green nodes indicate successful interventions, red nodes represent incidents without intervention, and yellow nodes correspond to false alarms.}
    \vspace{-10pt}
  \label{fig_1}
\end{wrapfigure}

These trade-offs underscore the need to couple early prediction with decision-making, often formalized as a two-stage framework known as predict-then-optimize~\cite{elmachtoub2022smart,shah2022decision,shao2017traveling,shao2020incorporating}. In this framework, a predictive model first analyzes the input features to generate forecasts, which are then used to define and solve a downstream problem that allocates limited resources based on model predictions. Building on this perspective, we revisit spatio-temporal prediction systems and analyze them through the lens of decision-making. Figure~\ref{fig_1} illustrates three common yet previously unformalized decision strategies: traditional, confidence-based, and adaptive. Traditional methods, often based on predictive models such as LSTM~\cite{memory1997sepp}, generate interventions after collecting complete spatio-temporal information. While accurate, such decisions are delayed, often missing early intervention windows. Moreover, acting on uncertain forecasts often results in high false alarms. To enable more robust decision-making, confidence-based approaches~\cite{wu2021quantifying} introduce decision rules that rely on the predicted probability of events. Actions are triggered only when confidence scores exceed predefined thresholds, thereby improving the likelihood of successful intervention. Another line of thought is based on early forecasting strategies~\cite{hartvigsen2020recurrent,shao2024stemo}, primarily designed to address the timeliness-accuracy trade-off. These adaptive methods dynamically adjust whether to initiate predictions based on contextual signals such as dynamic incident patterns. From the perspective of real-world utility, Confidence-based methods reduce false alarms, while adaptive strategies advance prediction timing, enabling timely intervention.

While some of these strategies offer unique advantages, they still face critical limitations in real-world settings. Most existing approaches rely on overly simplified assumptions regarding downstream resource availability, failing to account for the constrained and dynamic nature of environments such as emergency response or urban traffic management. For example, emergency response decisions must consider real-time ambulance availability in each region, yet unconstrained optimization often neglects these practical limitations, leading to infeasible actions. Moreover, in early spatio-temporal decision systems, the interactions between observations can be decomposed into multiple dimensions, including complex spatial-temporal dependencies, resource constraints, and their dynamic evolution over time. These multi-layered interactions demand a holistic perspective that jointly models the environment, resource availability, and decision impact. In this context, learning a practical Early Spatio-Temporal Decision Framework faces three key challenges: 1)~How to capture spatio-temporal dependencies under dynamic resource constraints, thereby enabling more accurate representations of evolving patterns. 2)~How to jointly model forecasting and decision-making such that predictions directly inform actionable interventions, enabling responses that maximize real-world utility in downstream tasks. 3)~How to define task-relevant metrics that jointly evaluate predictive accuracy and decision utility, ensuring a reflection of real-world performance in early spatio-temporal decision-making.

In this work, we proposed a novel framework, Adaptive Early Spatio-Temporal Decision Model ($\modelname$), which jointly models spatio-temporal prediction and downstream resource allocation in dynamic, resource-constrained environments. Specifically, we design a Resource-aware Spatio-Temporal interaction module (RaST) that incorporates a resource-adaptive graph learning layer to dynamically construct spatial connectivity based on real-time resource availability, effectively filtering out noisy or irrelevant interactions under resource constraints.
% , a dual-encoder architecture with dynamic spatio-temporal graph convolutional units. 
% The module applies a resource-adaptive graph learning layer to dynamically construct connectivity based on real-time resource availability, filtering out noisy or irrelevant spatial interactions under resource constraints. 
% The dual-encoder captures features that adapt to dynamic resource-aware prediction, while the dynamic resource availability is further mapped in the State Generator to produce fusion parameters that guide the integration of dual-encoder outputs. This module is responsible for disentangling complex interactions across different spatio-temporal and resource dimensions under dynamic environments. It captures long-term and short-term spatio-temporal dependencies. RaST not only enhances the expressiveness of the learned spatio-temporal representation but also generates representations that are directly actionable for downstream decision-making. After that, we introduce a preference-oriented decision agent (Poda). Poda transforms the predicted signals into resource-efficient and preference-aware intervention strategies, learning to optimize actions under resource constraints. 
The dual long- and short-term encoders within RaST capture features tailored to dynamic, resource-aware prediction.
These features are further processed by the State Generator to map real-time resource availability into fusion parameters that guide the integration of dual-encoder outputs. 
This design enables the model to disentangle complex interactions across spatio-temporal and resource dimensions in dynamic environments.
It effectively captures both long-term and short-term dependencies, resulting in expressive spatio-temporal representations that are directly actionable for downstream decision-making.
Building on RaST, we further introduce Poda, a preference-oriented decision agent. Poda transforms predicted signals into resource-efficient, preference-aware intervention strategies by learning to optimize actions under resource constraints.
The learning process in Poda is guided by explicit modeling of resource states and their dynamic transitions, enabling the agent to make resource-aware decisions under real-world constraints. By coupling state representation, action selection, and reward design with resource availability, Poda ensures effective and feasible interventions in dynamic environments. Additionally, Poda incorporates multi-objective reinforcement learning to balance multiple reward components such as intervention success rate and false alarm rate. By integrating a preference vector, Poda aligns decision-making with stakeholder-specific priorities or discovers hidden preferences to further enhance real-world decision-making performance.

Our contributions can be summarized as follows: 1) To the best of our knowledge, this is the first early spatio-temporal decision framework that directly optimizes downstream decision-making performance while producing actionable resource allocation decisions, bridging the gap between spatio-temporal learning and decision-making. 2) We propose a resource-aware spatio-temporal interaction module that explicitly models dynamic resource constraints during spatial and temporal encoding, enabling more context-aware and adaptable representations. 3) We establish a unified evaluation metric scheme tailored for early prediction and decision-making scenarios. This comprehensive metric scheme enables fine-grained analysis across timeliness, reliability, and efficiency, and goes beyond conventional accuracy-based metrics by capturing critical trade-offs faced in real-world resource allocation.
4) We conduct extensive experiments on real-world datasets, and the significant performance gains across various downstream baselines highlight the effectiveness of $\modelname$.

\section{Related Work}
\label{gen_inst}
\textbf{Spatio-Temporal Prediction.} 
Deep learning has advanced spatio-temporal prediction by modeling sequential and spatial dependencies. 
For example, GraphWaveNet~\cite{wu2019graph} and MTGNN~\cite{gao2022mtgnn} adopt Temporal Convolutional Networks (TCNs), while GMAN~\cite{zheng2020gman} and ASTGCN~\cite{guo2019attention} use self-attention to capture long-range temporal dependencies. 
AGCRN~\cite{bai2020adaptive} and DCRNN~\cite{li2017diffusion} integrate GRU-based architectures for dynamic environments.
STPGNN~\cite{kong2024spatio} introduces pivotal convolutions for localized pattern modeling, and PDFormer~\cite{jiang2023pdformer} leverages dynamic spatial-temporal attention with delay-aware messaging. 
Methods like STSGCN~\cite{song2020spatial}  and STG2Seq~\cite{bai2019stg2seq} enable synchronized learning of complex, localized interactions. Recent works, such as STEP~\cite{shao2022pre} and GPT-ST~\cite{li2023gpt}, adopt cutting-edge paradigms to improve spatio-temporal modeling. Despite these advances, accurately capturing informative spatio-temporal dependencies for downstream tasks remains a major challenge.

% Extensive research has focused on deep learning for spatio-temporal prediction. To model sequential patterns, GraphWaveNet\cite{wu2019graph} and MTGNN\cite{gao2022mtgnn} incorporate Temporal Convolutional Networks (TCNs). Simultaneously, GMAN\cite{zheng2020gman} and ASTGCN\cite{guo2019attention} employ self-attention networks to capture long-range temporal relations. For modeling spatio-temporal dependencies in dynamic environments, AGCRN\cite{bai2020adaptive} and DCRNN\cite{li2017diffusion} incorporate GRU-based coupled architectures. STPGNN\cite{kong2024spatio} introduces a pivotal convolution module, enabling localized modeling of spatio-temporal patterns. In parallel, PDFormer\cite{jiang2023pdformer} incorporates dynamic spatial-temporal attention with a delay-aware module to explicitly model message passing between regions. Additionally, methods such as STSGCN\cite{song2020spatial}  and STG2Seq\cite{bai2019stg2seq}  employ spatial-temporal synchronized learning to effectively capture complex and localized spatio-temporal interactions. Recent innovations have also explored the integration of spatio-temporal modeling with cutting-edge paradigms, such as STEP\cite{shao2022pre}  and GPT-ST\cite{li2023gpt} , to further enhance the understanding and modeling of spatial-temporal correlations. Despite these advancements, it remains challenging to accurately model the informative spatio-temporal dependencies that directly contribute to downstream task performance.

\textbf{Early Forecasting.}
Recent models for early forecasting aim to predict incomplete time series as early as possible while maintaining sufficient accuracy.
Popular approaches include shapelet-based~\cite{ghalwash2014utilizing,he2015early} and predictor-based methods~\cite{shao2023early,mori2017early,mori2017reliable,ghalwash2012early}.
To capture early discriminative patterns, DCRNN~\cite{li2017diffusion} incorporates diffusion convolution and ESTGCN~\cite{shao2023early} uses reinforcement learning to dynamically optimize decision timing.
STAEformer~\cite{liu2023spatio} models temporal dependencies via adaptive embeddings for early prediction from limited observations. 
Most of these methods focus on accuracy, instead of explicitly balancing accuracy and timeliness.
 Some approaches~\cite{mori2017early,mori2017reliable,tavenard2016cost} introduce tunable mechanisms to balance them. STEMO~\cite{shao2024stemo} uses reinforcement learning to model user preferences and optimize this trade-off. However, defining and balancing earliness and accuracy, along with establishing robust evaluation metrics, remains a key challenge in early forecasting research.

% In early forecasting, recent models aim to predict incomplete time series as early as possible while ensuring the required level of accuracy. Approaches such as shapelet-based methods \cite{ghalwash2014utilizing,he2015early} and predictor-based methods \cite{shao2023early,mori2017early,mori2017reliable,ghalwash2012early} have been widely adopted. To capture early discriminative patterns, DCRNN\cite{li2017diffusion} incorporates diffusion convolution mechanisms, while ESTGCN\cite{shao2023early} leverages reinforcement learning to dynamically determine optimal decision timing. STAEformer\cite{liu2023spatio} captures chronological dependencies via adaptive embeddings, enabling accurate early prediction from limited observations. Most of these methods primarily emphasize accuracy, instead of explicitly pursuing a balance between accuracy and timeliness. Some approaches\cite{mori2017early,mori2017reliable,tavenard2016cost} introduce mechanisms to adjust the trade-off between accuracy and timeliness through tunable parameters. To better quantify and optimise the balance between accuracy and timeliness, STEMO\cite{shao2024stemo} introduces a reinforcement learning framework that can infer hidden user preferences. However, the challenge of formally defining and balancing earliness and accuracy, as well as establishing appropriate evaluation metrics, remains a major bottleneck in early forecasting research.

\textbf{Optimization for Downstream Task.}
Downstream task optimization~\cite{wilder2019end,elmachtoub2022smart} plays a critical role in translating forecasting results into actionable decisions, particularly in domains such as transportation~\cite{shao2020incorporating,zheng2022supply,chen2024difflight} and medical care~\cite{hao2022reinforcement}.  Early approaches often relied on rule-based heuristics or classical optimization methods, such as Ant Colony Optimization and Genetic Algorithms, which require hand-crafted designs and lack adaptability across environments. 
Supervised learning techniques such as Pointer Networks~\cite{vinyals2015pointer} have been proposed to generate task-specific sequences under predefined constraints, showing effectiveness in structured decision problems. 
More recently, reinforcement learning has emerged as a flexible framework that enables agents to learn decision-making strategies through environment interaction. Existing  works~\cite{yang2019generalized,lin2022pareto,rame2023rewarded}  extend the capability to handle trade-offs among competing objectives and to discover Pareto-efficient policies under multi-objective settings.  Although these advancements are promising, most existing downstream task optimization approaches still treat prediction and decision-making as separate modules, limiting their ability to adapt under complex spatio-temporal dynamics and evolving downstream constraints.

\newtheorem{definition}{Definition}
% \section{Preliminary}
\section{Methodology}
\label{s3}
\subsection{Problem Formulation and Framework Overview}

\textbf{Problem Formulation.}
Consider a graph \( \mathcal{G} = (\mathcal{V}, \mathcal{E}) \), where \( \mathcal{V} = \{v_i\}_{i=1}^n \) denotes nodes and \( \mathcal{E} = \{e_{ij}\} \) denotes edges. Let \( \mathbf{X}_t = \{x_t^i\}_{i=1}^n \) denote node-level observations at time \( t \). Given the historical sequence \( \mathbf{X} = [\mathbf{X}_{t-L}, \ldots, \mathbf{X}_{t-1}, \mathbf{X}_t] \in \mathbb{R}^{L \times N \times C} \), a spatio-temporal model takes as input features \( \mathbf{X} \) and produces outputs \( \hat{\mathbf{X}}_T \). These outputs are then used to parameterize an optimization problem that is solved to yield actionable decisions. Specifically, spatio-temporal decision-making aims to select a binary action vector \( \mathbf{a} = \{a_i\}_{i=1}^n \in \{0,1\}^n \), where \( a_i = 1 \) indicates allocating a resource to node \( v_i \), subject to a resource constraint \( S_t \). Here, $S_t$ represents the available resources that remain unallocated at time $t$. The objective is to determine the best resource allocation strategy that maximizes decision utility while satisfying resource constraints:
% \begin{equation}
$
\mathbf{a}^* = \arg\max_{\mathbf{a} \in \{0,1\}^n} U\left( \mathbf{a} \mid \mathbf{X}, \mathcal{G} \right) \quad \text{s.t.} \quad \sum_{i=1}^n a_i \leq S_T
$.
% \end{equation}
The utility function is defined as \( U(\mathbf{a}) = \boldsymbol{\omega}^\top \mathbf{R}(\mathbf{a}) \), \( \mathbf{R}(\mathbf{a}) \in \mathbb{R}^d \) is a multi-objective reward vector that quantifies the decision-making performance such as success rate and false alarm rate. The preference vector \( \boldsymbol{\omega} \in \mathbb{R}^d \) encodes the relative importance of each component.

% the task is to predict future states \( \mathbf{Y} = [\mathbf{Y}_{t+1}, \ldots, \mathbf{Y}_{t+L'}] \in \mathbb{R}^{L' \times N \times C} \), where \( L' \) is the prediction horizon., the objective is to optimize the model parameters \( \theta \) by minimizing a loss function \( \mathcal{L}(\mathcal{M}_{\theta}(\mathbf{X}), \mathbf{Y}) \), which measures the discrepancy between the predicted and true future states. In spatio-temporal decision-making

\begin{figure}[t]
    \centering    \includegraphics[width=0.9\textwidth]{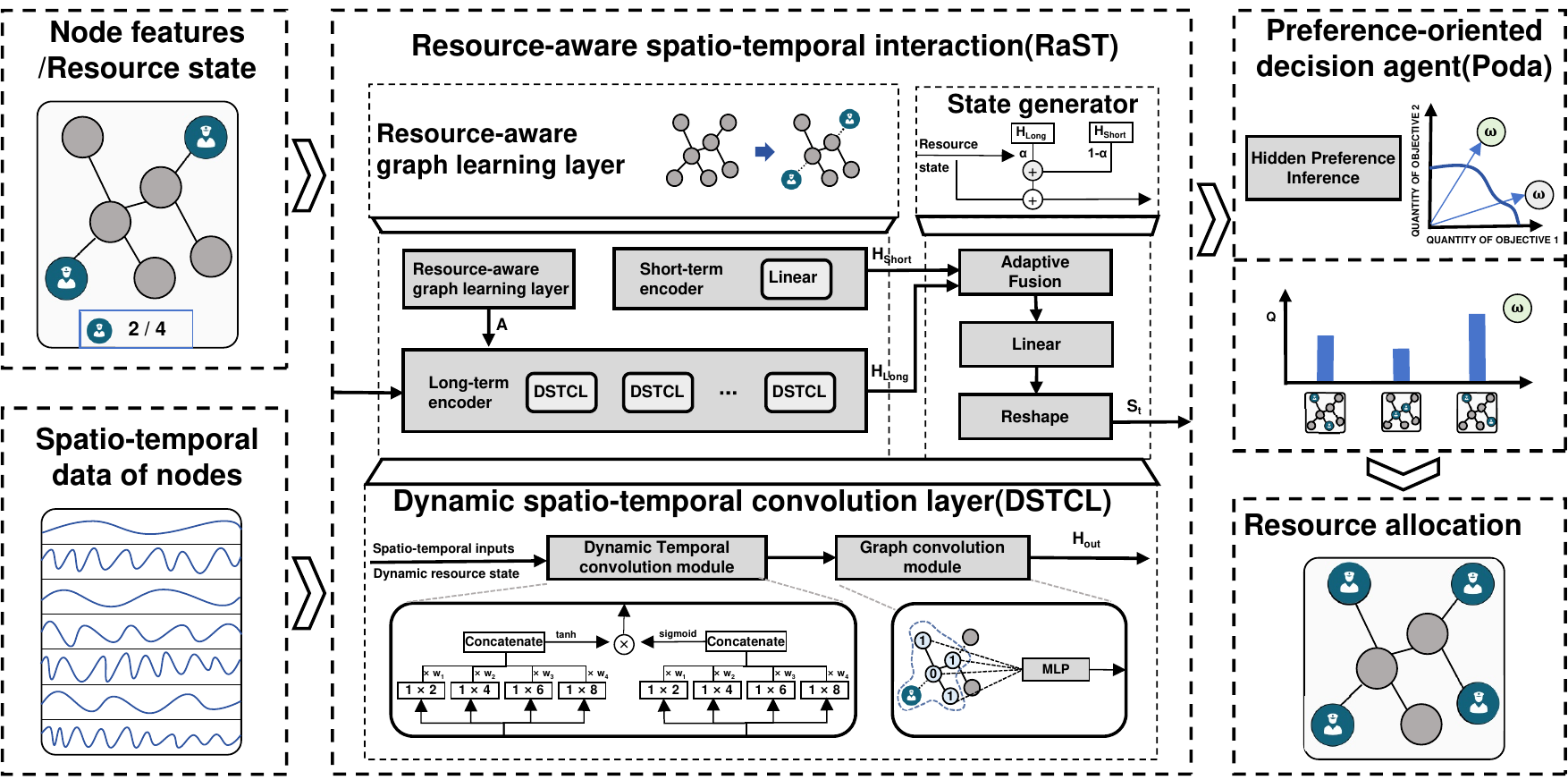}
    \caption{Framework overview of $\modelname$.}
    \label{fig:2}
\end{figure}

\textbf{Framework Overview.} 
The $\modelname$ is designed to advance spatio-temporal intelligence by synergistic integrating core components as illustrated in Figure \ref{fig:2}. We unify node-level resource features and dynamic spatio-temporal data into a harmonized format, and propose the RaST to capture long- and short-term dependencies under fluctuating resource constraints. The Poda infers hidden objectives and guides downstream decision-making by aligning multi-objective preferences with actionable resource allocation strategies, ensuring adaptability and optimality in resource-constrained environments.

\subsection{Resource-Aware Spatio-temporal Interaction Module (RaST)}
To better capture complex, multi-level interactions and incorporate resource states into spatio-temporal learning, we propose the RaST module. It leverages a resource-aware graph learning layer to perform message passing over nodes with unallocated resources, preserving useful structural information for decision-making. A state generator with an adaptive fusion mechanism, conditioned on real-time resource states, enables RaST to generate context-aware representations that support efficient and adaptive downstream decisions.

\textbf{Resource-Aware Graph Learning Layer}\quad
We introduce a resource-aware graph learning layer that adaptively constructs the adjacency matrix by explicitly incorporating the real-time availability of node-level resources. Unlike conventional methods~\cite{gao2022mtgnn} that compute a global, static adjacency solely from node embeddings, our approach integrates a resource mask to downweight or remove connections involving nodes currently engaged in active downstream tasks. This design enables the model to capture evolving spatial dependencies aligned with actionable nodes dynamically, enhancing decision relevance under resource constraints. It is illustrated as follows:
\begin{equation}
\begin{aligned}
& C_1 = \tanh(\beta Z_1 W_1), \quad C_2 = \tanh(\beta Z_2 W_2), 
\quad A^* = \mathrm{ReLU}\left( \tanh\left( \beta (C_1 C_2^\top - C_2 C_1^\top) \right) \right), \\
& M_{ij} =
\begin{cases}
1, & \text{if resource in both node } i \text{ and node } j \text{ are available}, \\
0, & \text{otherwise},
\end{cases} \\
% & A = A^* \odot M,
\end{aligned}
\end{equation}
where \( Z_1, Z_2 \) are learnable node embedding matrices, and \( W_1, W_2\) are their associated projection parameters. The matrices \( C_1 \) and \( C_2 \) encode projected node semantics, which are used to compute an asymmetric affinity matrix \( A^* \), modulated by a factor \( \beta \) to encourage directional interactions. A binary resource mask \( M \in \{0, 1\}^{N \times N} \) is then applied: $A = A^* \odot M$, allowing only idle node pairs to retain connectivity. This masking mechanism ensures that only nodes with available resources participate in graph message passing, effectively suppressing noisy or misleading signals from busy nodes. As a result, the final adjacency matrix \( A \) enables the model to capture resource-aware spatial-temporal dependencies for downstream decision-making.

\textbf{Dual Encoder}\quad 
To address the challenge of modeling spatio-temporal patterns under resource constraints, we design a dual-branch encoder architecture that capture long- and short- term information while retaining the desirable traits of each branch. In real-world scenarios, resource availability varies significantly across time steps and nodes. Uniformly extracting long-term dependencies can be computationally expensive and may introduce redundancy, whereas relying solely on short-term information may neglect crucial trends, leading to inaccurate forecasts. To resolve this, we decompose the temporal modeling process into two complementary branches. Specifically, the long-term branch employs Dynamic Spatio-Temporal Convolution Layers (DSTCL) to efficiently extract resource-aware dependencies, while the short-term branch captures immediate patterns. These branches dynamically fuse their representations conditioned on real-time resource states, as illustrated in Figure~\ref{fig:2}.

\textbf{Dynamic Spatio-Temporal Convolution Layer (DSTCL)} integrates temporal convolution with resource-aware graph convolution to explicitly model complex spatio-temporal patterns under resource constraints. To capture temporal dependencies of varying lengths while adapting to dynamic resource states, we propose a dynamically temporal convolution module. This module integrates the widely adopted dilated convolution strategy~\cite{yu2015multi} from graph neural networks and further introduces resource-aware gating factors, which dynamically adjust the temporal receptive field based on the available resource. We design a temporal inception layer consisting of four filter sizes, including $1 \times 2$, $1 \times 4$, $1 \times 6$, and $1 \times 8$. During the inception operation, larger convolutional filters are assigned higher fusion weights when resources are abundant, allowing the model to exploit long-range temporal patterns. Conversely, when resources are limited, smaller filters are prioritized to focus on short-term patterns. This resource-adaptive design ensures that the model allocates computational capacity efficiently by mapping the real-time resource state to the receptive field size, formulated as:
\begin{equation}
\mathbf{H_{in}} = \tanh(\operatorname{Inception}(\mathbf{X};\mathbf{W}_1, \mathbf{d}\mid \mathbf{S_T})) \odot \sigma(\operatorname{Inception}(\mathbf{X};\mathbf{W}_2, \mathbf{d}\mid \mathbf{S_T})),
\end{equation}
where $\mathbf{X}$ denotes input spatio-temporal features, $\mathbf{W}_1, \mathbf{W}_2$ are convolutional parameters, $d$ is the dynamically adjusted dilation factor. Inception applies parallel convolutions with multiple filter sizes and fuses their outputs using resource-adaptive fusion weights. Subsequently, a resource-aware adjacency structure $\tilde{\mathbf{A}} = \tilde{\mathbf{D}}^{-1}(\mathbf{A} + \mathbf{I})$ is dynamically learned by resource-aware graph learning layer. Finally, graph convolution is applied as:
% \begin{equation}
$\mathbf{H}_{\text{out}} = \text{MLP}\left( \tilde{\mathbf{A}} \mathbf{D}^{-1} \mathbf{H_{in}} \mathbf{W} \right)$
% \end{equation}
to selectively propagate information conditioned on real-time resource states. This design ensures that DSTCL captures accurate and resource-adaptive spatio-temporal representations.

\textbf{In the long-term encoder}, we extract long-range and dynamically evolving spatio-temporal dependencies by stacking multiple DSTCL layers.  This design allows the model to progressively integrate long-range dependencies while remaining sensitive to the current resource states. This branch can be formulated as:
% \begin{equation}
$\mathbf{H}_{\text{long}} = \text{DSTCL}_{\text{stack}}(\mathbf{X})$,
% \end{equation}
where $\text{DSTCL}_{\text{stack}}(\cdot)$ denotes the stacked dynamic spatio-temporal convolution layers. The output $\mathbf{H}_{\text{long}}$ retains long-range patterns while aligning with real-time resource conditions, and is used in subsequent fusion and decision modules.To stabilize training and preserve information flow, residual connections are applied between DSTCL layers, enabling the encoder to capture both original and transformed features effectively.

\textbf{In the short-term encoder}, we capture fine-grained and transient spatio-temporal patterns through a lightweight convolutional architecture. A 1D convolution first projects the input to a higher-dimensional temporal space to enhance feature expressiveness. Then, A spatio-temporal block (ST-Block), composed of residual temporal and spatial convolutions, further extracts short-term sequential and localized spatial dependencies. LayerNorm is applied to improve training stability and generalization. The process is formulated as:
$\mathbf{H}_{\text{short}} = \text{LayerNorm}(\text{ST-Block}(\text{Conv1D}(\mathbf{X})))$, where $\mathbf{H}_{\text{short}} $ is the final output encoding localized and timely features for downstream fusion. This compact design facilitates low-latency inference while preserving critical transient spatio-temporal information for downstream decision-making.

\textbf{State Generator}\quad 
To adaptively extract spatio-temporal information under dynamic resource constraints, we introduce a state generator that jointly performs feature fusion and prediction horizon selection. Guided by the principle that richer resources allow for more long-term information and longer prediction horizon, the module produces a resource-aware representation for decision-making. Given short-term and long-term encoded representations $\mathbf{H}_{\text{short}}$ and $\mathbf{H}_{\text{long}}$, we define resource-aware fusion ratio $\mathbf{\gamma} = \mathbf{S_t} / \mathbf{S}$, where $\mathbf{S_t}$ is the number of available resource at time $\mathbf{t}$, and $\mathbf{S}$ denotes the total available resource. We compute the fused representation as follows:
\begin{equation}
\begin{aligned}
\mathbf{H}_{\text{global}} &= (1 - \gamma) \cdot \mathbf{H}_{\text{short}} + \gamma \cdot \mathbf{H}_{\text{long}} \\
\mathbf{H}' &= \text{LayerNorm}(\mathbf{H}_{\text{global}} + \text{MHA}(\mathbf{H}_{\text{global}})) \\
\mathbf{H}_{\text{fused}} &= \text{LayerNorm}(\mathbf{H}' + \text{FFN}(\mathbf{H}')) 
\end{aligned}
\end{equation}

Here, $\mathrm{MHA}(\mathbf{X}) = \mathrm{Concat}_{h=1}^{\text{head}} \mathrm{Attention}(\mathbf{Q}_h, \mathbf{K}_h, \mathbf{V}_h) \mathbf{W}^O$ captures diverse contextual dependencies across nodes and time. The feed-forward network $\mathrm{FFN}(\mathbf{X}) = \max(0, \mathbf{X} \mathbf{W}_1 + \mathbf{b}_1) \mathbf{W}_2 + \mathbf{b}_2
$ performs non-linear transformation to enhance feature expressiveness. This refinement dynamically balances short- and long-term information according to the resource ratio $\gamma$. Based on this ratio, the state generator adaptively predicts the node-specific forecasting horizon $ \mathrm{round}(\mathbf{k}) \in \mathbb{Z}^n$, computed as $\mathbf{k} = \gamma \times \mathbf{k_{\max}}$, where $\mathbf{k_{\max}}$ is the predefined maximum prediction horizon. The fusion ratio $\gamma$ quantifies the degree of long-term information utilization at each node, providing an interpretable signal that guides downstream decision-making by explicitly incorporating node-specific temporal dependency patterns. Finally, by concatenating $\mathbf{H}_{\text{fused}}$ with real-time resource state $\mathbf{R}_{\text{state}}$ and node-specific location features $\mathbf{L}_{\text{node}}$, the state generator forms the complete state representation $\mathbf{H^T} = \,\mathbf{H}_{\text{fused}}\,\|\,\mathbf{R}_{\text{state}}\,\|\,\mathbf{L}_{\text{node}}\,$, aligning representation learning with resource-aware decision-making. Detailed configurations and implementation specifics are provided in Appendix~\ref{appendix:B.1}

\subsection{Preference-oriented Decision Agent (Poda)} 
\label{s3.3}
\textbf{Preference-aware Decision Module}\quad
To tackle the challenge of optimizing conflicting objectives in dynamic resource allocation, we propose a preference-oriented decision agent (Poda) that learns a Q-network capable of adapting to arbitrary preference vectors $\omega \in \Omega$. Instead of training one policy per preference, the agent learns a unified $Q_\theta: \mathcal{S} \times \mathcal{A} \times \Omega \rightarrow \mathbb{R}^d,
$ where $d$ is the number of objectives and $\mathcal{S}$ denotes the state space dynamically constructed from RaST. This integration ensures that the decision process is context-aware and preference-adaptive, rather than a decoupled post-processing step. At each time step $t$, the agent selects actions using a scalarized value $\omega^\top Q(s_t, a_t, \omega; \theta)$ under an $\epsilon$-greedy policy to avoid abundant exploitation:
\begin{equation}
a_t =
\begin{cases}
\arg\max_{a} \omega^\top Q(s_t, a, \omega; \theta), & \text{w.p. } 1 - \epsilon \\
\text{random action}, & \text{w.p. } \epsilon
\end{cases}
\end{equation}
where \(\theta \) denotes the trainable parameters of the Q-network, which are iteratively updated during training. The action $\mathbf{a}_t = \{ a_i^t \}_{i=1}^n$ is replaced by a random action with probability $\epsilon$, and $\epsilon$ decreases exponentially from $1$ to $0$ during training. For node $v_i$, $a_i^t = 1$ indicates allocating a resource at time $t$. The reward signal $r_t = \{ r_{t,\mathrm{acc}}^{i},\ r_{t,\mathrm{false}}^{i},\ r_{t,\mathrm{distance}}^{i},\ r_{t,\mathrm{time}}^{i} \}_{i=1}^{n}$ includes accuracy reward, false alarm reward, allocation distance reward, and temporal reward. The specific calculation is as follows:
\begin{equation}
r_t^i = \alpha \cdot r_{t,\mathrm{acc}}^{i} 
       - \beta \cdot r_{t,\mathrm{false}}^{i}
       - \gamma \cdot r_{t,\mathrm{distance}}^{i}
       + \delta \cdot \ r_{t,\mathrm{time}}^{i},
\label{eq:reward}
\end{equation}
where $\alpha$, $\beta$, $\gamma$, and $\delta$ are trade-off coefficients controlling the importance of each objective. Accuracy reward measures whether the allocated actions correctly match actual events. False alarm reward encourages responsible decision-making by encouraging the agent to avoid resource allocation to irrelevant nodes. Allocation distance reward captures the overall cost of resource reallocation, promoting cost-effective dispatch decisions. Temporal reward captures execution timeliness by rewarding earlier successful predictions, motivating the agent to make timely interventions with minimal execution delays. The agent is trained via two losses: a vector regression loss to accurately estimate per-objective Q-values and an auxiliary scalarization loss to align preference-weighted targets:

\begin{equation}
\begin{aligned}
\mathcal{L}_A(\theta) &= \mathbb{E} \left[ \left\| y - Q(s_t, a_t, \omega; \theta) \right\|_2^2 \right],
\quad y = r_t + \gamma \max_{a, \omega'} \omega^\top Q(s_{t+1}, a, \omega'; \theta),
\\
\mathcal{L}_B(\theta) &= \mathbb{E} \left[ \left| \omega^\top y - \omega^\top Q(s_t, a_t, \omega; \theta) \right| \right].
\end{aligned}
\end{equation}
The overall loss is then a weighted sum: 
% \begin{equation}
$\mathcal{L}(\theta) = (1 - \lambda)\mathcal{L}_A + \lambda\mathcal{L}_B$,
% \end{equation}
where $\lambda$ is annealed from 0 to 0.6 during training to ensure stable optimization. This design ensures that the agent progressively transitions from general multi-objective learning to preference-aligned policy optimization, which better reflects task-specific trade-offs and decision requirements.

\textbf{Inference of Hidden Preferences}\quad
In many real-world scenarios, user preferences over objectives are implicit, making it essential to infer hidden preferences from behavioral signals. Inspired by~\cite{yang2019generalized}, we treat the preference vector $\omega$ as a latent variable and aim to estimate its value from interactions. For each trajectory, we optimize $\omega$ to maximize the expected scalarized return:
% \begin{equation}
$\omega^* = \arg\max_{\omega \in \Omega} \sum_{t} \omega^\top r_t$.
% \end{equation}
This inferred $\omega^*$ is then used to guide action selection and policy updates, enabling the agent to adapt its behavior even in the absence of explicit user guidance.By jointly learning a preference-parameterized Q-function and continuously refining latent preferences, the proposed agent achieves robust decision-making across diverse tasks and objective trade-offs.

\textbf{Resource-constrained Environment}\quad
To simulate realistic resource-constrained decision-making, we implement a discrete-time environment that maintains and updates a dynamic allocation of limited resources across $N$ nodes. At each timestep, a fixed number of resources are available, and each resource can be in one of two states: idle or allocated. Cooldown mechanisms reflect delayed recovery of deployed resources, introducing temporal constraints on reuse. Given an allocation action vector $a_t \in \{0,1\}^N$, the environment uses the Hungarian algorithm to solve a minimum-cost matching problem between supply nodes (those with available resources) and demand nodes (those selected by $a_t$). The reallocation incurs a cost based on a predefined distance matrix $D \in \mathbb{R}^{N \times N}$. This setup enforces spatial and temporal constraints, ensuring that the action strikes a balance between timely interventions and efficient resource utilization. More details of the methodology and training algorithm can be found in the Appendix~\ref{appendix:B.2}.
\section{Experiment}
\label{s4}
\subsection{Experimental Setup}
\label{s4.1}
We evaluate our model under a resource-constrained environment, unlike existing work that often
% . This is in contrast to prior works, which typically 
ignore such limitations.
% , leading to unrealistic assessments of downstream performance. 
As shown in~\cite{shah2022decision, kotary2021end}, 
% Studies~\cite{shah2022decision, kotary2021end} have shown that 
optimization quality varies substantially when task-specific constraints are introduced, as the structure of decisions becomes highly sensitive to prediction errors under such limitations. 
Hence, we introduce authentic spatio-temporal resource constraints and report evaluations across a specific time horizon, aligning model assessment with practical settings. 

% \textbf{Datasets and Baselines} \quad 
%  We conducted experiments on four real-world datasets: 1)~\textbf{NYC}\footnote{https://www.kaggle.com/datasets/mysarahmadbhat/nyc-traffic-accidents}: The traffic accident dataset used in this study consists of incident records collected in New York City. 2)~\textbf{NYPD}\footnote{https://data.cityofnewyork.us/Public-Safety/NYPD-Complaint-Data-Historic/qgea-i56i}: The crime dataset used in this study is a collection of data from the New York Police Department. 3)~\textbf{EMS}\footnote{https://data.cityofnewyork.us/Public-Safety/EMS-Incident-Dispatch-Data/76xm-jjuj}: The emergency dataset used in this study consists of incident allocation records collected by the New York Fire Department. 4)~\textbf{XTraffic}\cite{gou2024xtraffic}: The XTraffic dataset contains traffic sensor data collected from Orange. We compare $\modelname$ with a range of baselines that cover recent advances in both spatio-temporal forecasting and early prediction, including HA~\cite{liu2004summary}, 
% LSTM~\cite{memory1997sepp}, 
% ARIMA~\cite{box2015time}, 
% DCRNN~\cite{li2017diffusion}, GraphWaveNet~\cite{wu2019graph},  STAEformer~\cite{liu2023spatio},
% EARLIEST~\cite{hartvigsen2019adaptive},
% ESTGCN~\cite{shao2023early} and
% STEMO~\cite{shao2024stemo}.  Please refer to Appendix~\ref{appendix:C} for more details about these datasets and methods.

\textbf{Datasets and Baselines} \quad 
 We conducted experiments on four real-world datasets: 1)~\textbf{NYC}~\cite{nyc_opendata}: The traffic accident dataset consists of incident records collected in New York City. 2)~\textbf{NYPD}~\cite{nypd_data}: The crime dataset is a collection of data from the New York Police Department. 3)~\textbf{EMS}~\cite{ems_data}: The emergency dataset consists of incident allocation records collected by the New York Fire Department. 4)~\textbf{XTraffic}~\cite{gou2024xtraffic}: The XTraffic dataset contains traffic sensor data collected from Orange. We compare $\modelname$ with a range of baselines that cover recent advances in both spatio-temporal forecasting and early prediction, including HA~\cite{liu2004summary}, 
LSTM~\cite{memory1997sepp}, 
ARIMA~\cite{box2015time}, 
DCRNN~\cite{li2017diffusion}, GraphWaveNet~\cite{wu2019graph},  STAEformer~\cite{liu2023spatio},
EARLIEST~\cite{hartvigsen2019adaptive},
ESTGCN~\cite{shao2023early} and
STEMO~\cite{shao2024stemo}.  Please refer to Appendix~\ref{appendix:C} for more details about these datasets and methods.

\textbf{Metrics} \quad 
We design six metrics detailed as follows.
% spanning three key aspects.  Specifically, 
To measure decision effectiveness, we report the Success Rate \textbf{(SR)} and False Alarm Rate \textbf{(FAR)}, which quantify the accuracy and reliability of interventions. To assess spatio-temporal efficiency, we include Average Distance \textbf{(AD)} and Average Early Prediction Time \textbf{(AET)}, reflecting the spatial cost and temporal lead of response. To evaluate resource utilization, we also adopt the Resource Utilization Rate \textbf{(RUR)} and Cost-Effectiveness Ratio \textbf{(CER)}, which characterize the benefit-cost trade-off under limited resource conditions. 
\begin{itemize}[leftmargin=*, align=parleft, labelsep=1em]
\item Success Rate \textbf{(SR)}: Measures the proportion of successful interventions overall true events, defined as $\text{SR} = \frac{1}{n} \sum_{i=1}^n \mathbb{1}[ \text{succ}(a_i, \hat{y}_i) ]$, where $\text{succ}(a_i, \hat{y}_i) = 1$ if event occurs and $a_i = 1$.
\item  False Alarm Rate \textbf{(FAR)}: Captures the ratio of false alarms among all positive responses through $\text{FAR} = \frac{1}{n} \sum_{i=1}^n \frac{\mathbb{1}\left[ \text{false}(a_i, \hat{y}_i) \right]}{\mathbb{1}\left[ \text{succ}(a_i, \hat{y}_i) \right] + \mathbb{1}\left[ \text{false}(a_i, \hat{y}_i) \right]}$. 
\item  Average Distance \textbf{(AD)}: Computes the average spatial cost of allocated resources, $\text{AD} = \frac{1}{n} \sum_{i=1}^n a_i \cdot d_i$, where $d_i$ is the denotes the travel cost to node $v_i$. 
\item  Average Early Prediction Time \textbf{(AET)}: Measures how early interventions occur, $\text{AET} = \frac{1}{n} \sum_{i=1}^n \mathbb{1}[a_i = 1 \wedge\hat{y}_i = 1] \cdot \Delta t_i$, $\Delta t_i$ is time from resource allocation to event occurrence. 
\item  Resource Utilization Rate \textbf{(RUR)}: Reflects average usage of resources, $\text{RUR} = \frac{1}{n} \sum_{i=1}^n \frac{ \text{sum}(a_i)}{S}$, where $S$ is the total resources.
\item   Cost-Effectiveness Ratio \textbf{(CER)}: Quantifies average reward per unit resource allocation cost, $\text{CER} = \frac{1}{n} \sum_{i=1}^n \frac{r_i}{c_i}$, where $r_i$ and $c_i$ denote reward and allocation cost.
\end{itemize}

\begin{table*}[t]
    \centering
    \small
    \setlength{\tabcolsep}{6pt} % 调整列间距
    \renewcommand{\arraystretch}{0.9} % 调整行间距
    \caption{Results comparison on tasks benchmark. Best results are \textbf{bold}. We report the mean for three trials.}
    \vspace{-4pt}
    \label{tab:comparison}
    \resizebox{\textwidth}{!}{
    \begin{tabular}{l|l|l|l|l|l|l|l|l|l|l|l}
        \toprule
        \textbf{Dataset} & \textbf{Metric} & \textbf{HA} & \textbf{ARIMA} & \textbf{LSTM} & \textbf{DCRNN} & \textbf{GWNet} & \textbf{STAEformer}& \textbf{EARLIEST}& \textbf{ESTGCN}& \textbf{STEMO} & 
        \textbf{ASTER}\\
        \midrule
        \multirow{6}{*}{NYC}
         & SR $\uparrow$ & 0.24 & 0.25 &  0.28 & 0.09 & 0.05 & 0.27 & 0.26 & 0.23 &   0.11 & \textbf{0.36} \\    
         & FAR $\downarrow$ & 0.80 & 0.78 & 0.75 & 0.93 & 0.95 & 0.77 & 0.91 & 0.92 & 0.86   & \textbf{0.65} \\
        & AD $\downarrow$ & 305.56 & 385.18 & 175.32 & 159.64 & 947.97 & \textbf{120.65} & 233.68 & 231.65 &  140.39 & 148.35 \\
        & AET $\uparrow$  & 1.45 & 1.48 & 1.47 & 1.43 & 2.46 & 2.26 & 5.38 & \textbf{6.76} & 5.77  & 3.26 \\
        & RUR $\downarrow$ & 0.20 & 0.20 & 0.20 & 0.20 & 0.20 & 0.20 & 0.50 & 0.50 &  0.50 &  \textbf{0.16} \\
        & CER $\uparrow$ & 0.03 & 0.15 & 0.07 & 0.01 & 0.07 & 0.11 & 0.02 & 0.22 &  0.01  &  \textbf{0.24} \\
        \midrule
        \multirow{6}{*}{NYPD}
         & SR $\uparrow$ & 0.28  & 0.14 & 0.10 & 0.26 & 0.28 & 0.29 & 0.29 & 0.29 & 0.27  & \textbf{0.35} \\    
         & FAR $\downarrow$ & 0.24 & 0.07 & \textbf{0.02} & 0.28 & 0.23 & 0.22 & 0.48 & 0.47 & 0.50 & 0.25 \\
        & AD $\downarrow$ & 95.25  & 47.77 & 76.04 & 53.08 & 74.75  & 44.64 & 88.54 & 62.56 & 103.21 & \textbf{41.64} \\
        & AET $\uparrow$ & 1.29 & 1.00 & 1.01 & 1.27 & 1.28  & 1.28 & 7.14 & 6.40 & \textbf{12.95} & 3.32  \\
        & RUR $\downarrow$ & 0.34  & 0.34 & 0.34 & 0.34 &  0.34  &  0.34 & 0.50 & 0.50 & 0.50 & \textbf{0.18}  \\
        & CER $\uparrow$ &  0.31 & 0.42 & 0.26 & 0.55 & 0.41  & 0.77 & 0.25 & 0.35 &  0.32& \textbf{0.94} \\
        \midrule
        \multirow{6}{*}{EMS}
         & SR $\uparrow$ & 0.17  & 0.14 & 0.20 & 0.20 & 0.23 & 0.22 & 0.25 & 0.18 & 0.22 & \textbf{0.28}  \\    
         & FAR $\downarrow$ & 0.36  & \textbf{0.04} & 0.06 & 0.25 & 0.18 & 0.16 & 0.63 & 0.74 & 0.67 &0.16  \\
        & AD $\downarrow$ & 136.99 & 267.98 & 103.48 & 155.23 & 142.73 & 98.20 & 157.79 & 150.13 & 185.09 & \textbf{89.96} \\
        & AET $\uparrow$ & 1.34  & 1.58 & 1.53 & 1.84 & 1.79 & 1.79 &  6.13  & 6.87 & \textbf{7.69} &3.82 \\
        & RUR $\downarrow$ & 0.20 & 0.20 & 0.20 & 0.20 &  0.20 & 0.20 & 0.50 & 0.50 & 0.50 &\textbf{0.16}  \\
        & CER $\uparrow$ & 0.32 & 0.04 & 0.51 & 0.33 & 0.40 & 0.59  & 0.17 & 0.12 & 0.13 &\textbf{0.68}   \\
        \midrule
        \multirow{6}{*}{XTraffic}
         & SR $\uparrow$ & 0.18  & 0.16 & 0.24 & 0.16 & 0.19 & \textbf{0.34} & 0.31 & 0.16 & 0.30 & 0.30  \\    
         & FAR $\downarrow$ & 0.71 & 0.75 & 0.89 & 0.92 & 0.91 & 0.83 & 0.90 & 0.88 & 0.92 & \textbf{0.68} \\
        & AD $\downarrow$ & 35.74 & 56.32 & 38.52 & 29.81 & 38.95 & 41.11 & 31.39 & \textbf{26.31} & 72.11 & 43.82 \\
        & AET $\uparrow$ & 1.72 & 2.71 & 2.46 & 1.82 & 1.72 & 1.65 & 3.56 & 6.26 &  \textbf{7.08} & 3.35 \\
        & RUR $\downarrow$ & 0.25 & 0.25 & 0.25 & 0.25 & 0.25 & 0.25 & 0.50 & 0.50 & 0.50 & \textbf{0.22} \\
        & CER $\uparrow$ & 0.24 & 0.18 & 0.13 & 0.08 & 0.07 & 0.12 & 0.10 & 0.02 & 0.06  & \textbf{0.28} \\
        \bottomrule
    \end{tabular}}
\end{table*}
\subsection{Performance Comparison}
1)~\textbf{Performance comparison among baselines.} In Table~\ref{tab:comparison}, We evaluated the performance of $\modelname$ against other methods across multiple spatio-temporal datasets. $\modelname$ consistently outperforms all baseline methods across diverse datasets and evaluation metrics. It achieves an average 11.15\% improvement in success rate over the second-best approaches and demonstrates comprehensive superiority in downstream utility metrics such as RUR and CER.\begin{wraptable}{r}{0.57\textwidth}
    \centering
    \vspace{-10pt}
    \caption{Performance against different resource levels.}
    \label{tab_p2}
    \resizebox{\linewidth}{!}{
    \begin{tabular}{cccccccccc}
        \toprule
        \multirow{3}{*}{\textbf{Model}} & \multicolumn{9}{c}{\textbf{EMS}} \\  % 只跨数据列
        \cmidrule(lr){2-10} &
        \multicolumn{3}{c}{\textbf{Low Resources}} & \multicolumn{3}{c}{\textbf{Medium Resources}} & \multicolumn{3}{c}{\textbf{High Resources}} \\
         & \textbf{SR} $\uparrow$& \textbf{RUR}$\downarrow$  & \textbf{CER}$\uparrow$& \textbf{SR}$\uparrow$ &\textbf{RUR} $\downarrow$& \textbf{CER}$\uparrow$ & \textbf{SR} $\uparrow$& \textbf{RUR}$\downarrow$ & \textbf{CER} $\uparrow$\\
        \midrule
        \textbf{GWNeT} & 0.11 & 0.20 & 0.45 & 0.23 & 0.20 & 0.40 & 0.32 & 0.20 & 0.35 \\
        \textbf{STAEformer}  &0.09  & 0.20  & 0.35 & 0.22 & 0.20 & 0.59 & 0.25 & 0.20 & 0.57 \\
        \textbf{STEMO}  & 0.12 & 0.50 & 0.11 & 0.22  &  0.50& 0.13& 0.31& 0.50 & 0.18  \\
        \midrule
        \textbf{ASTER}     & \textbf{0.13}  & \textbf{0.16}  & \textbf{0.48} & \textbf{0.28}& \textbf{0.16} & \textbf{0.68} & \textbf{0.34} & \textbf{0.16} & \textbf{0.66} \\
        \bottomrule
    \end{tabular}
    }
    \vspace{-10pt}
\end{wraptable} These results highlight the effectiveness of integrating forecasting with decision-making under dynamic resource constraints, enabling $\modelname$ to maximize intervention success and resource efficiency by allocating limited resources to the right locations at the right time. 2)~\textbf{Robustness in different resource constraints.} We constructed scenarios with varying levels of resource constraints to evaluate the robustness of $\modelname$ in realistic resource allocation settings. Specifically, we simulated limited-resource scenarios by restricting the number of total resources between consecutive resource allocations. Experimental results summarized in Table~\ref{tab_p2} reveal that $\modelname$ consistently maintains superior performance across these constrained settings. This demonstrates that our framework relaxes the dependency on abundant resources by promoting adaptive and resource-aware decision-making, ensuring performance even under different resource constrain.

\subsection{Ablation Study}
\begin{wrapfigure}{r}{0.52\textwidth}
  \centering
    \vspace{-10pt}
  \includegraphics[width=0.51\textwidth]{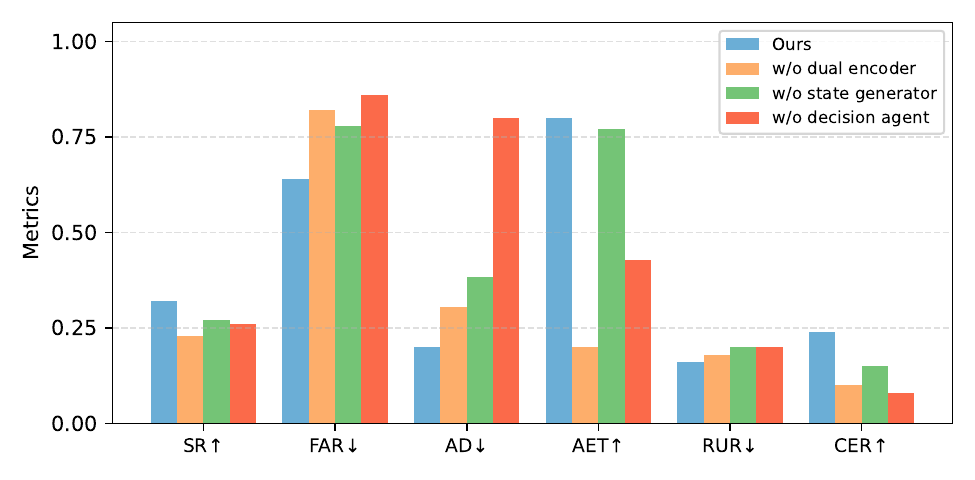}
  \caption{Ablation studies of $\modelname$ on NYPD, AD and AET are linearly rescaled.}
  \label{fig:abalation}
  \vspace{-15pt}
\end{wrapfigure}

To evaluate the contribution of each core component in $\modelname$ and its effectiveness in decision making, we design a set of ablation variants: 1)~\textbf{w/o dual encoder}: remove the long- and short-term encoder and replace it with a GCN-based encoder, 2)~\textbf{w/o state generator}: disable the fusion of long- and short-term encoders by replacing the learned fusion weights with static averaging, 3)~\textbf{w/o decision agent}: replace the Poda with a vanilla DQN agent using scalarized rewards. As shown in Figure~\ref{fig:abalation}, removing any core component leads to performance drops across key metrics. The results of "w/o dual encoder" indicate that capturing resource information along with the corresponding spatio-temporal dependencies is essential for supporting effective downstream actions. The performance of "w/o state generator" further highlights the effectiveness of dynamically fusing long- and short-term representations based on real-time resource states. Finally, the outcomes of "w/o decision agent" demonstrate that removing multi-objective decision-making leads to noticeably inferior results across all evaluation metrics.

\subsection{Case Study}
% 1)~\textbf{Resource-aware allocation behavior.} 

\begin{figure}[H]
  \centering
    \vspace{-10pt}
  \begin{subfigure}[b]{0.49\linewidth}
    \centering \includegraphics[width=\linewidth]{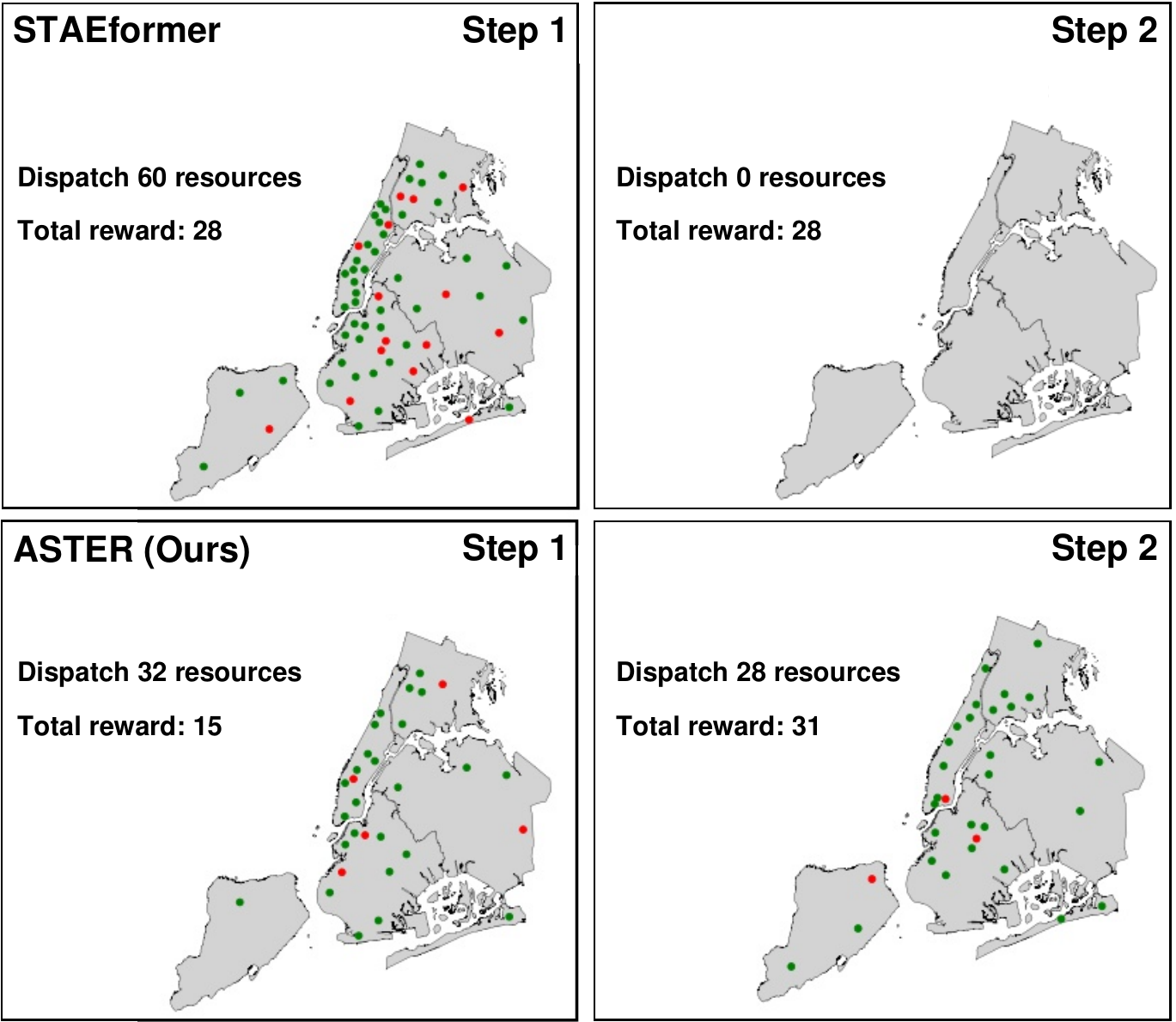}
    \caption{Resource-aware allocation behavior.}
 \label{fig:case_a}
  \end{subfigure}
  \hfill
  \begin{subfigure}[b]{0.425\linewidth}
    \centering  \includegraphics[width=\linewidth]{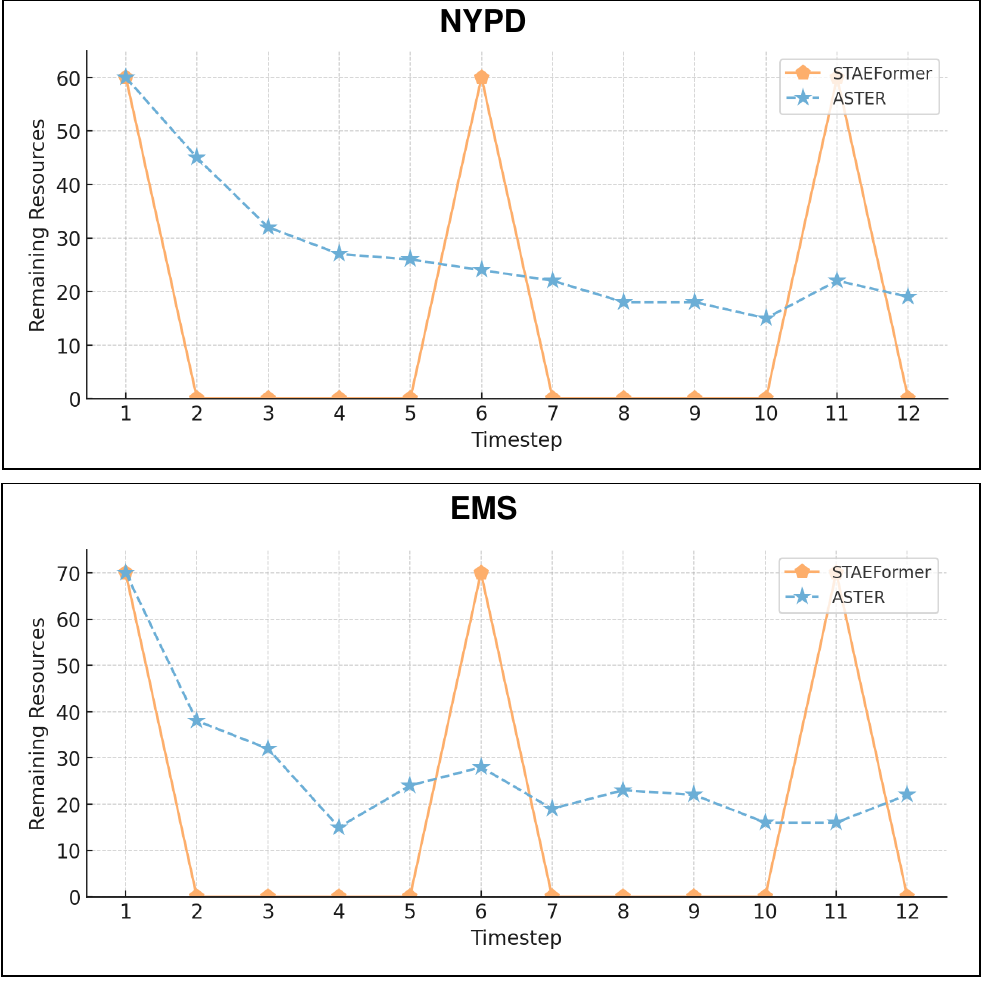}
    \caption{Long-term resource level.}
\label{fig:case_b}
  \end{subfigure}
  \caption{Case studies of ASTER on NYPD and EMS dataset.}
  \vspace{-10pt}
\end{figure}

In Figure~\ref{fig:case_a}, we present the resource allocation patterns of STAEformer and $\modelname$ over 2 consecutive time steps in the NYPD validation set. STAEformer with probabilistic scheduling tends to allocate all resources in the initial stage, leaving the system unable to respond to subsequent demands, whereas $\modelname$ continuously evaluates the remaining resources at each decision step and adaptively adjusts the planning horizon, enabling the agent to naturally balance long- and short-term utility. 
% 2)~\textbf{Long term resource level.} 
Figure~\ref{fig:case_b} illustrates the long-term resource retention patterns on the NYPD and EMS datasets. Compared to STAEFormer, which tends to exhaust resources prematurely, $\modelname$ maintains a more balanced usage over time. This behavior demonstrates $\modelname$'s ability to incorporate resource-awareness into decision-making, while adapting to dynamic task demands across different domains.

\section{Conclusion and Discussion}
\label{s5}
In this work, we present $\modelname$, a unified model that integrates spatio-temporal learning and decision-making to address their traditional separation. $\modelname$ captures spatio-temporal dependencies via resource-aware graph learning and adaptively fuses long- and short-term features based on real-time resource availability. A preference-oriented decision agent employs multi-objective reinforcement learning to align actions with task-specific trade-offs. By jointly optimizing early prediction and downstream decisions, $\modelname$ bridges the gap between forecasting and decision-making. $\modelname$ contributes to the development of adaptive, resource-efficient AI systems that can support timely and effective interventions in high-stakes scenarios. Its ability to reduce unnecessary interventions and optimize resource usage promotes responsible AI deployment, helping cities and communities better manage limited resources while minimizing risk and loss to the public. More detailed discussions on limitations and potential extensions are provided in the Appendix~\ref{appendix:D.1}.

% \section{Limitation and Future Work}

\bibliographystyle{unsrt}

\bibliography{references}

%%%%%%%%%%%%%%%%%%%%%%%%%%%%%%%%%%%%%%%%%%%%%%%%%%%%%%%%%%%%

%%%%%%%%%%%%%%%%%%%%%%%%%%%%%%%%%%%%%%%%%%%%%%%%%%%%%%%%%%%%

\newpage
\appendix
\noindent\rule{\textwidth}{0.8pt}
\vspace{-5mm}
\begin{center}
    { \Large \bfseries \textbf{Supplementary Material for ASTER}}\\[1.5em]
    \vspace{-4mm}
\end{center}
\vspace{-4mm}
\noindent\rule{\textwidth}{0.4pt}
\section{Explanation of Relevant Concepts}
\label{appendix:A}

\subsection{The Concept and Scope of Spatio-Temporal Early Decision-Making}
\label{appendix:A.1}
In our study, various decision signals are collected from spatio-temporal sources reflecting the states of real-world systems. For instance, a spatio-temporal early decision model typically integrates incident records, node-level contextual information, and dynamic resource states. We collect and organize decision signals from multiple spatio-temporal sources into a representation. The goal of our work is to explore the extraction of spatio-temporal dependencies under resource constraints and to enable node-level decision-making within a unified model. To this end, the concept of decision-making here is to jointly model representation learning and resource allocation in an integrated system. Therefore, our work does not target at regression or classification problems, but proposes an integrated model that directly connects spatio-temporal dynamics with actionable resource-aware decisions, thus getting rid of gap between prediction and decision-making. While our experiments focus on resource allocation in emergency early response scenarios, the proposed model is general and can be extended to other real-world decision-making tasks involving dynamic constraints.

\subsection{Dynamic Resource Allocation}
Dynamic resource allocation is a common operational problem in many real-world systems such as emergency response, transportation logistics, healthcare services, and urban management. These systems are characterized by the need to continuously reassign limited resources (e.g., ambulances, police units) to dynamically emerging demands across spatial regions. In such scenarios, spatial costs arise from the physical movement of resources, typically measured by distance. Meanwhile, resource constraints such as an ambulance being occupied on an ongoing mission, prevent its immediate reuse, limiting its availability for future demands. This type of problem often involves matching available resources to target demands in a utility-driven manner, which can be formulated as a maximum-utility matching problem under resource constraints.

\section{Methodology Details}\label{app:dataset}
\label{appendix:B}
\subsection{Illustration of Data Representation and Fusion}
\label{appendix:B.1}

\begin{figure}[H]
    \centering 
    \vspace{-3mm}
    \includegraphics[width=0.9\textwidth]{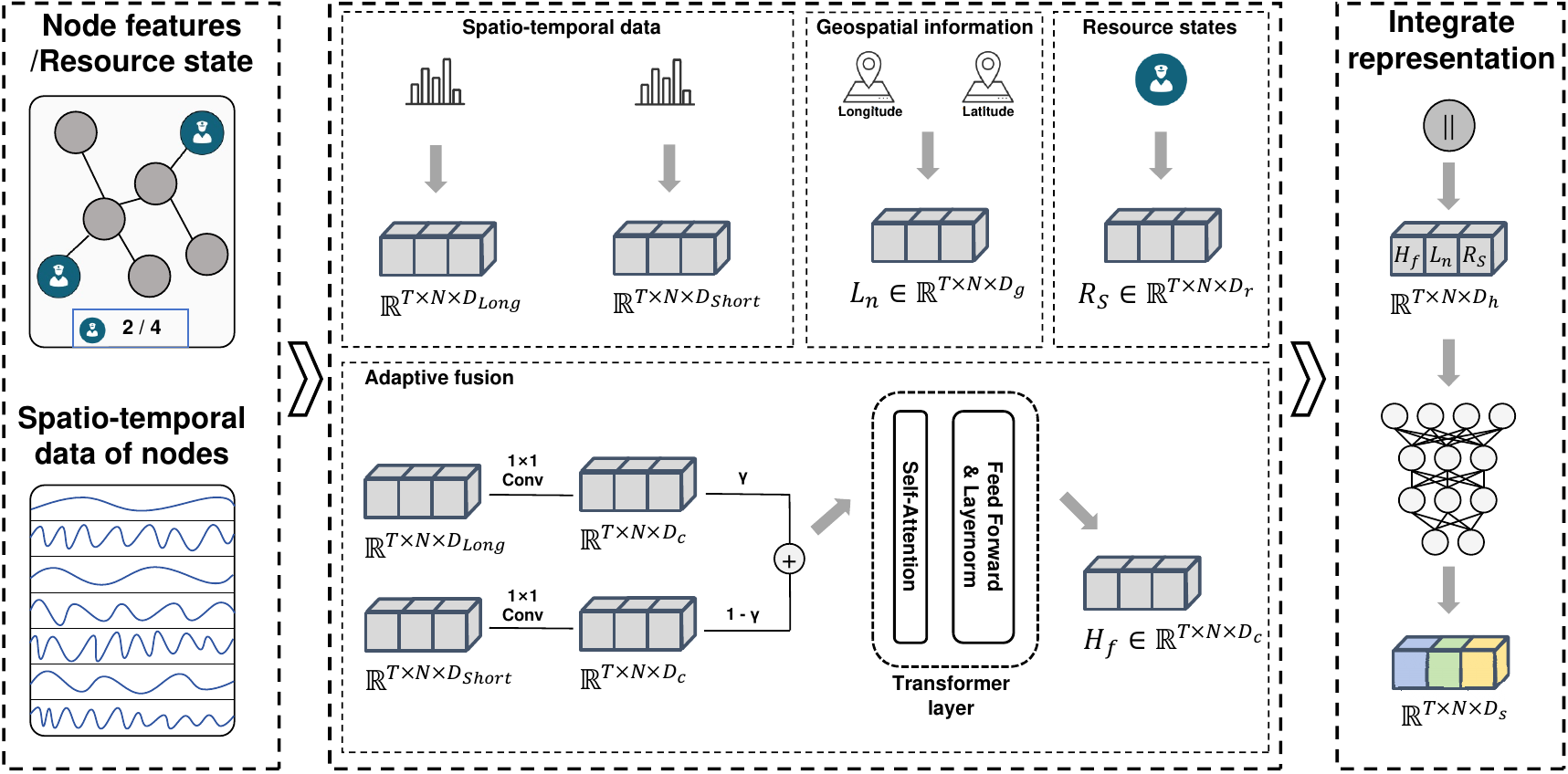}
    \caption{This figure illustrates the process of Spatio-Temporal Representation and Integration within the  ASTER. $T$ and $N$ denote the length of the time series and the number of nodes, respectively. The diagram details how multi-source signals, including spatio-temporal features and dynamic resource states, are jointly encoded across different dimensions, with the right corner aggregating these encodings into a resource-aware spatio-temporal representation for downstream decision-making.}
    \label{sp_1}
\end{figure}

\subsection{Training Protocol}
\label{appendix:B.2}
The overall model is trained end-to-end using alternating updates for the RaST and the Poda. At each training step, the RaST first estimates future event representations over $k$ steps. These representations are then used to construct a resource-aware state vector $s_t$, which fuses encoded spatio-temporal features with current environment conditions (i.e., resource states and node locations). Based on $s_t$ and the sampled preference vector $\omega$, the agent selects an action $a_t$ to allocate resources. Once actions are executed in the environment, updated resource states and the resulting reward vector $r_t$ (composed of accuracy reward, false alarm reward, allocation distance reward, and temporal reward) are recorded. Transitions $(s_t, a_t, r_t, s_{t+1})$ are stored in a replay buffer. The Q-network is trained via multi-objective temporal-difference learning using both standard Bellman error and scalarization-aligned loss. Meanwhile, the RaST is updated via a masked loss that only accounts for prediction errors before the dynamically predicted step $k$. This joint training paradigm tightly integrates upstream forecasting with downstream decision-making. Unlike conventional cascaded baselines that passively act on fixed predictions, our model dynamically modulates the prediction horizon and resource allocation via preference- and context-aware policies. The complete algorithmic workflow of ASTER is detailed in Algorithm~\ref{ag_1}.

\begin{algorithm}[ht]
\caption{Training of ASTER}
\label{ag_1}
\begin{algorithmic}[1]
\Require Spatio-temporal data $X_t$, location $L_t$, preference sampling distribution $\mathcal{D}$, path $p_\lambda$ for weight $\lambda$ increasing from $0$ to $0.6$, encoder, network $\mathcal{Q}$, resource environment, maximum epoch number $E$, batch size b, maximum prediction horizon $k_m$, resource level.

\Statex \hspace{-1.5em}\textbf{Parameter}: Corresponding learnable parameters $\theta_e$, $\theta_d$, and $\theta$.

\Statex \hspace{-1.5em}\textbf{Output}: Learned model

\State Initialise encoder, decoder, network $\mathcal{Q}$, replay buffer $\mathcal{D}_\mathcal{T}$
\For{$e = 1$ to $E$}
    \State Initialise resource environment based on resource level.
    \State Sample a batch from $X_t$
%    \State Get current resource states $\text{R}_\text{state}$ from resource environment
        \For{sample $i = 1$ to $b$}
        \State Observe recoded values $X_i$ and resource state $R_t$ from resource environment
        \State $H_t,k \gets \text{Encoder}(X_t, R_t, k_m ; \theta_e)$
        \State  Construct state $s_t$ from hidden state $H_t$, resource state $R_t$, and location features $L_t$.
        \State $a_t \gets 
        \begin{cases}
        \arg\max_a \omega^\top \mathcal{Q}(s_t, a, \omega; \theta), & \text{w.p. } 1 - \epsilon \\
        \text{random}, & \text{w.p. } \epsilon
        \end{cases}$    
        \State $\hat{X}_T \gets \text{Decoder}(H_t, k ;\theta_d)$
        \State Update resource environment after action $a_t$
        \State Receive vectorized reward $r_t$ and observe $s_{t+1}$.
        \State Store transition $(s_t, a_t, r_t, s_{t+1})$ in $\mathcal{D}_\mathcal{T}$.
        \If{update network}
            \State Sample $N_\tau$ transitions $(s_j, a_j, r_j, s_{j+1}) \sim \mathcal{D}_\mathcal{T}$.
            \State Sample $N_\omega$ preferences $\{\omega_i \sim \mathcal{D}\}$
            \State Compute $y_{ij}$ for all $1 \leq i \leq N_\omega$, $1 \le j \le N_\tau$. according to Equations 6
            \State Update $\theta$ by descending its stochastic gradient $\nabla_\theta L(\theta)$ according to Equations 6
            \State Increase $\lambda$ along the path $p_\lambda$
        \EndIf
        %\State Compute predictor loss $\mathcal{L}_\theta$ using masked MSE over $k_i$ steps
        \State Update $\theta_e$ and $\theta_d$ by descending its stochastic gradient according to MSE($X_T$,$\hat{X}_T$)
        \EndFor
\EndFor

\Statex \hspace{-1.5em} \textbf{Return} learned model
\end{algorithmic}
\end{algorithm}

%=====

\section{Additional Experiment Details}
\label{appendix:C}

\subsection{Dataset Details}
\label{appendix:C.1}
We provide detailed descriptions of the datasets used in this study, including the number of records in the original data, the number of regions into which the data was divided, and the time intervals. Details are presented in Table~\ref{c.1}. The specific data preprocessing is as follows:

\textbf{NYC}: We collect NYC traffic accident records~\cite{nyc_opendata} from January to August 2020. Each record provides information such as accident time and location. We filter out records with missing values or invalid coordinates beyond the NYC boundary. Then we divide the city into 225 grids covering the main urban areas. Accident counts are aggregated in each grid into 1-hour intervals, representing the number of accidents occurring in that region and time slot. We further incorporate time of day (tod) and day of week (dow) as temporal context, resulting in a structured spatio-temporal dataset with input features [value, tod, dow] for  spatio-tempora early decision-making tasks.

\textbf{NYPD}: We collect NYPD crime complaint records~\cite{nypd_data} from January 2014 to December 2015. Each record includes information such as the crime occurrence time, location, and administrative regions. We clean the dataset by removing invalid or incomplete records. The dataset is divided into 77 regions based on administrative regions. Crime incidents in each region are aggregated into 1-hour intervals, reflecting the number of reported crimes per hour in that area. We further add time of day (tod) and day of week (dow) as temporal context, resulting in a structured spatio-temporal dataset with input features [value, tod, dow] for downstream crime prediction and resource allocation tasks.

\textbf{EMS}: We collect EMS dispatch records~\cite{ems_data} from January 2011 to November 2011. Each record contains the dispatch time, location, and the corresponding postal code region. We clean the dataset by removing invalid or incomplete records. The city is divided into 231 postal code regions, with emergency incidents aggregated into 30-minute intervals, capturing the number of EMS dispatches per region and time slot. We further incorporate time of day (tod) and day of week (dow) as temporal context, resulting in a structured spatio-temporal dataset with input features [value, tod, dow] for downstream emergency demand forecasting and resource allocation tasks.

\textbf{XTraffic}: We collect traffic sensor data and incident records from XTraffic~\cite{gou2024xtraffic} from January 2023 to December 2023. Following the processing strategy of the authors, we treat the nearest sensor to each incident as a node. Each incident is matched to its nearest node based on road network metadata. Then we select data within Orange and surrounding areas, divided into 91 sensor-defined regions, with traffic incidents aggregated into 2-hour intervals, reflecting the number of incidents captured by each sensor in the corresponding time slot. We further add time of day (tod) and day of week (dow) as temporal context, resulting in a structured spatio-temporal dataset with input features [value, tod, dow] for downstream traffic incident prediction and resource allocation tasks.

\textbf{In addition to accident records}, we simulate operational constraints by defining a limited number of deployable resources, such as a fixed number of ambulances (e.g., 50 units), based on real-world emergency service capacities. This constraint introduces a realistic setting for evaluating resource allocation strategies under capacity limits. 

Specifically, we categorize low, medium, and high resource levels based on the ratio between available resources $S$ and the estimated per-step resource demand. 1) Low resource level: Resources are fewer than demands, requiring strict prioritization. 2) Medium resource level: Resources roughly match demands, reflecting balanced operational conditions. 3) High resource level: Resources exceed demands, allowing broader coverage but risking more false alarm. These settings reflect typical operational scenarios ranging from resource scarcity to over-provisioning, and help evaluate model robustness under varying levels of pressure.

\begin{table}[ht]
\centering
\caption{The detailed information of the four datasets.}
\label{c.1}
\resizebox{0.9\textwidth}{!}{
\begin{tabular}{lccccl}
\toprule
\textbf{Dataset} & \textbf{Time Span} & \textbf{\#Regions} & \textbf{\#Time Steps} & \textbf{Time Interval} & \textbf{Resource Levels} \\
\midrule
NYC      & \begin{tabular}[c]{@{}c@{}}01/01/2020-\\ 31/08/2020\end{tabular} & 225 & 5808  & 1h      & \begin{tabular}[l]{@{}l@{}}Low :  $S \leq 53$ \\ Medium : $ 53 < S\leq88$\\ High : $S > 88$\end{tabular} \\
\midrule
NYPD     & \begin{tabular}[c]{@{}c@{}}01/01/2014-\\ 31/12/2015\end{tabular} & 77  & 17521 & 1h      & \begin{tabular}[l]{@{}l@{}}Low : $S \leq 30$\\ Medium : $ 30 < S\leq50$\\ High : $S > 50$ \end{tabular} \\
\midrule
EMS      & \begin{tabular}[c]{@{}c@{}}01/01/2011-\\ 30/11/2011\end{tabular} & 231 & 15986 & 30 min  & \begin{tabular}[l]{@{}l@{}}Low : $S \leq 86$\\ Medium : $ 86 < S\leq144$\\ High : $S > 144$\end{tabular} \\
\midrule
XTraffic & \begin{tabular}[c]{@{}c@{}}01/01/2023-\\ 31/12/2023\end{tabular} & 91  & 4380  & 2h      & \begin{tabular}[l]{@{}l@{}}Low : $S \leq 22$   \\ Medium : $ 22 < S\leq 38$ \\ High : $S > 38$\end{tabular} \\
\bottomrule
\end{tabular}}
\end{table}

\subsection{Implementation and Evaluation  Details}
\label{appendix:C.2}
\textbf{Implementation details}\quad We partitioned each dataset into training, validation, and testing sets using a 7:1:2 ratio. \modelname{} generates the next dynamic k time steps actions on the past 168 observations. All data were normalized to zero mean and unit variance. For the RaST module, the temporal convolution layers used four inception filter sizes $\{1 \times 2, 1 \times 4, 1 \times 6, 1 \times 8\}$ with resource-adaptive fusion weights. The hidden dimension for the state representation was set to 64. The Poda agent was trained with a vectorized Q-network outputting 4-dimensional multi-objective rewards. We adopted the adam optimizer with an initial learning rate of $1 \times 10^{-3}$. Our model was implemented with PyTorch on a Linux system  equipped with Tesla V100 32GB.

\textbf{Baseline  details}\quad 
To assess the decision-making effectiveness of ASTER, we evaluate it against a comprehensive suite of nine baseline models categorized into three groups: 1) traditional forecasting methods~(HA~\cite{liu2004summary}, 
LSTM~\cite{memory1997sepp}, 
ARIMA~\cite{box2015time}), 2) spatio-temporal forecasting models~(DCRNN~\cite{li2017diffusion}, GraphWaveNet~\cite{wu2019graph},  STAEformer~\cite{liu2023spatio}), 3) early forecasting approaches~(EARLIEST~\cite{hartvigsen2019adaptive},
ESTGCN~\cite{shao2023early}, STEMO~\cite{shao2024stemo}). To ensure a fair comparison, we apply a unified downstream resource allocation strategy across all baseline methods. Specifically, we rank the regions based on the average of their predicted event intensities over a multi-step horizon, and allocate limited resources in descending order of these scores, subject to available resource constraints. This simulates realistic deployment with these models under resource constraints. By standardizing the allocation mechanism, we ensure that differences in decision-making outcomes arise solely from the forecasting quality, enabling consistent and fair evaluation across models. The average results are reported in our Table~\ref{tab:comparison} in the main paper.

\textbf{For traditional models}, we consider Historical Average (HA), ARIMA, and LSTM as representative baselines. These models capture temporal patterns based on statistical heuristics or sequential modeling without incorporating spatial dependencies. Despite their simplicity, they provide strong baselines for evaluating how much spatio-temporal and early prediction mechanisms contribute to downstream decision quality. 

\textbf{For spatio-temporal forecasting models}, we include DCRNN, GraphWaveNet, and STAEformer as strong baselines that explicitly model spatial and temporal dependencies across regions. These models leverage graph-based architectures or attention mechanisms to capture complex spatio-temporal correlations in dynamic environments. They have been widely adopted for fine-grained forecasting tasks and provide a meaningful benchmark for evaluating the benefit of early decision-making capabilities. All spatio-temporal models are implemented and trained under unified framework, from which we extract their optimal parameters and subsequently evaluate them within our downstream decision-making environment.

\textbf{For early forecasting approaches}, we adopt EARLIEST, ESTGCN, and STEMO as representative models that explicitly aim to make predictions as early as possible while maintaining sufficient accuracy. These models are designed to balance the trade-off between timeliness and reliability, often incorporating dynamic halting mechanisms to determine the optimal prediction point. They align closely with our problem setting, where early and effective decision-making is crucial under resource constraints. All early forecasting models are evaluated using their recommended configurations and integrated into our unified assessment framework to ensure consistent downstream evaluation.

\textbf{Evaluation details}\quad 
For fairness, all models are evaluated under the same weekly sliding window setup and downstream resource allocation policy. For each node, we compute a scalar score (e.g., the average predicted values across the future horizon). Based on this score and the number of currently available resources, we identify a subset of target nodes that require intervention in the current decision step. Once the set of selected target nodes is determined, a minimum-cost matching algorithm (e.g., the Hungarian algorithm) is applied to assign the available resources to these nodes, minimizing the total spatial allocation cost. This setup reflects practical dispatch scenarios, where decisions are made based on predicted urgency and subject to both spatial constraints and resource availability. We simulate resource allocation using a discrete-time environment that tracks the state of each node’s resources and cooldowns. The continuous timeline is divided into weekly intervals and a set of evaluation metrics are computed at each interval, including Success Rate (SR), False Alarm Rate (FAR), Average Distance (AD), Average Early Time (AET), Resource Utilization Rate (RUR), and Cost-Effectiveness Ratio (CER). For robustness, we report the average metric values across all weekly test intervals.

\subsection{Parameter Sensitivity Analysis}
\label{appendix:C.3}
\begin{wrapfigure}{r}{0.30\textwidth}
    \vspace{-5mm}
    \setlength{\columnsep}{3pt}
    \begin{subfigure}{\linewidth}
        \centering        \includegraphics[width=\linewidth]{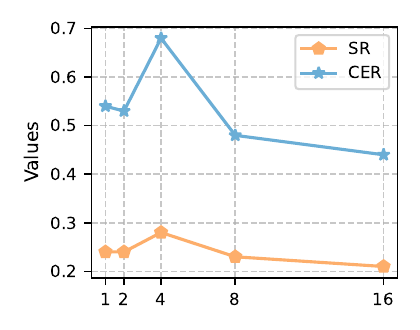}
            \vspace{-5mm}
        \caption{Number of Heads}
    \end{subfigure}
    
    \begin{subfigure}{\linewidth}
        \centering        \includegraphics[width=\linewidth]{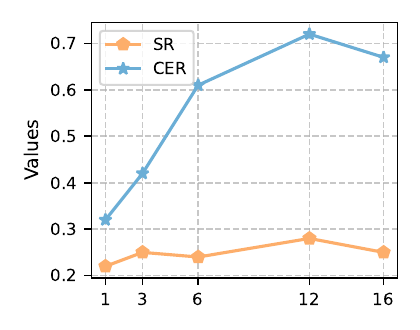}
            \vspace{-5mm}
        \caption{Maximum Prediction Horizon}
    \end{subfigure}
    
    \caption{Parameter
    Sensitivity Analysis of ASTER.}
    \label{fig:mask_ratio_analysis}
    \vspace{-5mm}
\end{wrapfigure}

To examine the impact of different hyperparameters on ASTER, we conduct parameter experiments on the EMS dataset. The results of these experiments are depicted in Figure~\ref{fig:mask_ratio_analysis}. We report how changes in hyperparameters affect SR (Success Rate) and CER (Cost-Effectiveness Ratio). SR reflects the model’s ability to make successful interventions, while CER captures the overall efficiency of downstream resource utilization. In our investigation, we specifically focus on the number of attention heads in RaST and the predefined maximum prediction horizon as the main parameters of interest. Based on the findings presented in Figure 8, we observe that the optimal effect is achieved when the number of attention heads is set to 4. Increasing the number of attention heads beyond this value does not improve the model’s ability to learn better representations; instead, it leads to a decline in decision performance. Furthermore, we explore the impact of the predefined maximum prediction horizon, which determines the temporal scope over which the model is expected to make forward-looking decisions. This parameter effectively governs how far into the future the model can forecast and, consequently, how much uncertainty is introduced during training. Performance improves when the prediction horizon is moderately extended, enabling earlier interventions and improved accuracy of critical events. The best performance is obtained when the maximum prediction horizon is set to 12. However, excessive extension of this horizon amplifies temporal noise and degrades performance across key metrics, particularly the Cost-Effectiveness Ratio. These results suggest that an appropriately chosen prediction horizon not only improves the model's ability to balance earliness and accuracy but also ensures stable learning under dynamic resource constraints.

\subsection{Revealing Hidden Preference}
\label{appendix:C.4}
To assess the agent’s capability to adapt to unknown preferences, we conduct experiments on the EMS dataset. During the training phase, the agent has no access to the true preference and instead learns a general-purpose policy. In the adaptation phase, we provide the agent with tasks associated with hidden one-hot preferences and evaluate its ability to learned hidden preferences.\begin{wraptable}{r}{0.34\textwidth}
\centering
\small
\renewcommand{\arraystretch}{1}
\vspace{-0.5em}
\resizebox{\linewidth}{!}{
\begin{tabular}{lcccc}
\toprule
         & \textbf{Accuracy} & \textbf{False alarm} & \textbf{Allocation distance} & \textbf{Temporal} \\
\midrule
\textbf{v1} & \cellcolor{highlight}1 & 0 & 0 & 0 \\
\textbf{v2} & 0 & 0 & 0 & \cellcolor{highlight}1 \\
\bottomrule
\end{tabular}
}
\caption{Inferred preferences of the agent on different tasks.}
\label{tab:wrap_v1v2}
\end{wraptable} Specifically, we test two task settings (v1 and v2), where the hidden preferences are [1, 0, 0, 0] and [0, 0, 0, 1], corresponding to accuracy and timeliness~(as described in Section~\ref{s3.3}), respectively. As shown in Table~\ref{tab:wrap_v1v2}, the agent successfully infers and selects preferences that closely match the hidden objectives, indicating effective alignment with the true underlying priorities.

\section{Others}
\label{appendix:D}
\subsection{Limitations and Future Work}
\label{appendix:D.1}
$\modelname$ aims to align early spatio-temporal learning with downstream decision-making under dynamic resource constraints. However, the current $\modelname$ model faces limitations in terms of the interpretability of its learned spatio-temporal representations. Additionally, real-world environments often impose more complex and dynamically evolving constraints, which the current model does not fully accommodate. To address these limitations, future research may explore the development of more interpretable spatio-temporal graph learning methods to uncover the underlying dynamics of spatio-temporal interactions. Furthermore, we plan to investigate dynamic reasoning mechanisms that can adapt to real-time fluctuations in resource availability, enabling more robust and responsive decision-making.
Moreover, extending $\modelname$ to support decentralized or multi-agent coordination represents a natural next step, particularly for large-scale or partially observable settings. We also aim to explore the broader applicability of the early decision paradigm in domains such as environmental monitoring and urban planning, where timely and resource-efficient interventions are critical.

\end{document}